\documentclass{svproc}

\usepackage{url}

\usepackage{wrapfig}
\usepackage{graphicx}
\usepackage[sectionbib,numbers]{natbib}
\usepackage{multirow}

\usepackage{textcomp}
\usepackage{microtype}
\usepackage{booktabs}

\begin{document}
\mainmatter 
             
\title{Deployment of Customized Deep Learning based Video Analytics On Surveillance Cameras}
\titlerunning{Customized Deep Learning based Video Analytics} 
\author{Pratik Dubal\thanks{Equal Contribution} \and Rohan Mahadev\footnotemark[1] \and Suraj Kothawade\footnotemark[1] \and \\ Kunal Dargan \and Rishabh Iyer}
\authorrunning{Pratik Dubal et al.}
\tocauthor{Pratik Dubal, Rohan Mahadev, Suraj Kothawade, Kunal Dargan, Rishabh Iyer}

\institute{AitoeLabs (www.aitoelabs.com)}

\maketitle 

\begin{abstract}
This paper demonstrates the effectiveness of our customized deep learning based video analytics system in various applications focused on security, safety, customer analytics and process compliance. We describe our video analytics system comprising of Search, Summarize, Statistics and real-time alerting, and outline its building blocks. These building blocks include object detection, tracking, face detection and recognition, human and face sub-attribute analytics. In each case, we demonstrate how custom models trained using data from the deployment scenarios provide considerably superior accuracies than off-the-shelf models. Towards this end, we describe our data processing and model training pipeline, which can train and fine-tune models from videos with a quick turnaround time. Finally, since most of these models are deployed on-site, it is important to have resource constrained models which do not require GPUs. We demonstrate how we custom train resource constrained models and deploy them on embedded devices without significant loss in accuracy. To our knowledge, this is the first work which provides a comprehensive evaluation of different deep learning models on various real-world customer deployment scenarios of surveillance video analytics.
By sharing our implementation details and the experiences learned from deploying customized deep learning models for various customers, we hope that customized deep learning based video analytics is widely incorporated in commercial products around the world.
\keywords{Deep Learning, Convolutional Neural Networks, Computer Vision, Customized Video Analytics}
\end{abstract}
\section{Introduction}
Visual Data, in the form of images, videos and live streams, has been growing at an unprecedented rate in the last few years. While this massive amount data is a blessing for Data Science, as it helps in improving the predictive accuracy, it is also a curse since humans are unable to consume this large amount of data. Moreover, today, machine-generated videos (via Drones, Dash-cams, Body-cams, Surveillance cameras etc.) are being generated at a rate higher than what we as humans can process. Among machine-generated videos, surveillance videos are one of the largest contributors to this growth. Surveillance cameras are deployed in several verticals, including office facilities, road intersections for traffic monitoring, ATMs and Banks, Hospitals, Manufacturing Facilities, Industrial Plants, Construction Sites, Educational Institutions, Retail stores and Malls, Hotels and Restaurants etc. Each of these verticals have their own unique video analytics applications. In most scenarios, video analytics is used for security purposes (detecting loitering and intrusion, asset tampering, suspicious activity or object detection). In other scenarios, video analytics is used for process compliance, e.g. if an event in a manufacturing plant has happened on time, or whether it was done as desired. In retail scenarios and hotels, the information from video analytics is used for getting insights in customer pattern (e.g. heat-map, flow-map, counts, dwell-times etc.) While all these applications sound very different, the analytics building blocks are the same.

\begin{wrapfigure}{R}{0.5\textwidth}
\centering
\includegraphics[width=0.5\textwidth]{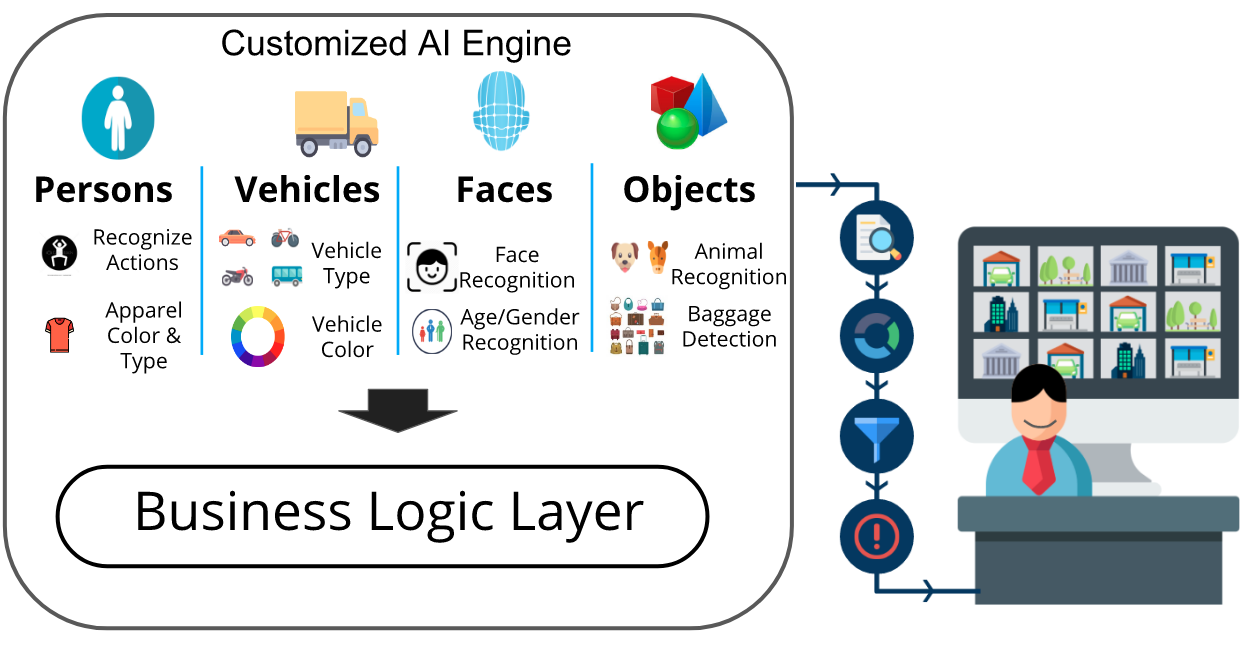}
\caption{End-to-End process for analytics}
\label{newton}
\end{wrapfigure}

Figure~\ref{newton} demonstrates the process clearly. The analytics engine consists of several building blocks, including object detection, tracking, face and human detection, human and face sub-attribute recognition, vehicle detection and vehicle sub-attribute recognition etc. The information from the analytics engine is then passed on to a business logic layer, which applies rules based on the analytics output. For example, using human detection (localizing where a human is in the video frame) if a human enters a demarcated area, it sends out a real-time alert. Similarly, by tracking the paths of the human in the video, we can compute the heat-map and flow-map of human movement.

The following sections outline the advancement of deep learning in computer vision, followed by the recent advances and challenges of video analytics for surveillance applications. Finally, we outline the main contributions of this paper.

\subsection{Advancement of Deep Learning in Computer Vision}
\label{subSec:advDLCV}
Current approaches to all but a few Computer Vision tasks involve the use of Deep Convolutional Neural Networks (CNNs). CNNs generated a lot of interest after the successful performance of AlexNet~\cite{AlexNet} in the ImageNet Large Scale Visual Recognition Challenge (ILSVRC) 2012 competition~\cite{ILSVRC15}. Following its triumph, there was an upsurge in the number of deep CNN models that were being used across the Computer Vision community. The winner of ILSVRC 2014 was the even deeper GoogLeNet, which was the first CNN model to have a fundamentally different architecture than AlexNet. It was followed by ResNet~\cite{Resnet}, the winner of ILSVRC 2015, which was an astonishing 152 layers deep. It won the competition by achieving an error rate of 3.57\%, beating humans at the image classification task.

Similarly, for Object Detection tasks, there has been a significant advancement in the use of CNNs in the last lustrum. It started with the introduction of the Region based family of networks~\cite{girshick2014rich, girshick2015fast,ren2015faster}. Recently, the Single Pass family of networks, consisting of YOLO and Tiny Yolo~\cite{redmon2016you,yolov3}, along with the Single Shot MultiBox Detector (SSD)~\cite{SSD} have emerged as the state-of-the-art models in object detection.
Their extremely fast inference times allow for object detection to take place in real-time, thus broadening the areas of application where Deep Learning can be used for vision tasks.

While there has been a remarkable improvement in the performance of Object Detection models for real-time tasks, an important role is still played by Object Tracking algorithms. Tracking algorithms are much faster than detection algorithms and help preserve the identity of an object being tracked when detection fails. Object Tracking is highly dependent on the quality of object detections. With good detections, the performance of a simple tracking algorithm increases drastically.

State-of-the-art face recognition techniques such as DeepFace~\cite{DeepFace} and Deep Face Recognition~\cite{Parkhi15} both consist of CNNs. Some of the highest results on the 'Labeled Faces in the Wild' (LFW) dataset~\cite{LFWTech} have been achieved by supervised CNNs~\cite{Learned-Miller2016}. Recently, a ResNet based face embeddings model has been proposed by~\citeauthor{dlibFace}~\cite{dlibFace}. In almost all comparisons, deep face recognition models have outperformed older hand-crafted face embedding models.

\subsection{Challenges of Video Analytics in Security and Surveillance}
A lot of research has gone into video analytics systems for security and surveillance applications. Over the past two decades, video analytics companies have been providing solutions and products for video analytics in several domains. ~\citeauthor{gong2011security}~\cite{gong2011security} and~\citeauthor{gouaillier2009intelligent}~\cite{gouaillier2009intelligent} provide a good summary of the technology, problems, as well as the companies which are building analytics products in this space. Most surveillance cameras have a fixed angle of view, and for this reason, video analytics on these surveillance cameras are slightly easier than other forms of video analytics on moving cameras. Many video analytics problems can therefore be solved by background subtraction algorithms, which essentially use motion information to generate contours and motion blobs. ~\citeauthor{sobral2014comprehensive}~\cite{sobral2014comprehensive} provide a very comprehensive survey of background subtraction algorithms for motion analytics. These algorithms work well in low traffic situations, and where one wants high sensitive alerts for problems like intrusion detection, motion detection and asset tampering. However, background subtraction algorithms are mostly unsupervised algorithms and are not trained to specifically detect humans or other objects of interest. As a result, they cannot distinguish between motion caused by shadows or leaf movements, viz-a-viz a human or animal intrusion, and often generate a lot of false alarms. However, background subtraction algorithms are extremely fast and scale very well in embedded applications. Due to privacy and bandwidth issues, it is often not feasible and prohibitively costly to deploy video analytics solutions on the cloud. As a result, it is essential to develop resource constrained video analytics solutions which can be deployed on premise. Deep learning has dominated the landscape of computer vision for the past few years, and almost all video analytics applications can be solved with high accuracies via deep learning. However, deep learning algorithms are resource hungry and require expensive GPU cloud servers to deploy. Given this, developing resource constrained, locally deployable and embedded deep learning solutions is critical. Several very recent advances like the MobileNet family of models~\cite{howard2017mobilenets}, X-NOR networks~\cite{rastegari2016xnor} and Tiny-YOLO~\cite{yolov3} have enabled deployment of resource constrained on embedded devices. 

\subsection{Our Contributions}
The following are the main contributions of this paper:
\begin{itemize}
\item A systematic overview of what it takes to build an end-to-end video analytics system for the surveillance domain.
\item A data collection and training pipeline to ensure fast turnaround times, along with data sampling and augmentation tricks used.
\item A comprehensive data analysis, in terms of the number of images used for training in each deployment scenario, and other subtle tricks and lessons learned to get these to work.
\item A comprehensive evaluation of how custom models based on deployment scenarios provide considerably superior accuracies than off-the-shelf models. 
\item Demonstrate how the powerful deep learning models can be run at reasonable frame-rates on edge devices. We also compare accuracies of resource constrained models viz-a-viz cloud enabled models. 
\item Lastly, we show how the resource constrained edge models perform considerably better than off-the-shelf GPU enabled models, thereby emphasizing the power of model customization for deployments. 
\end{itemize}
To our knowledge, this paper provides the first comprehensive evaluation of various computer vision tasks such as object detection and localization, face detection and face recognition, face and human sub-attribute recognition etc. In each case, we provide comprehensive evaluation of deep learning models on real-world customer data and deployment scenarios.

\section{Video Analytics System Overview} \label{system-overview}
To have a robust video analytics system in the surveillance domain, a pivotal role is played by the accuracy of models and the inference times. The occurrence of false positives in detections and the delay in transmission of real-time alerts may potentially hinder the effective utilization of the system. Thus, our research emphasizes on the creation of deep learning based models which are capable of achieving high accuracy rates, without compromising on the inference times.

Building on top of the recent advancements in deep learning, we propose our multi-faceted video analytics system, which focuses on performing real-time analytics such as object detection, face analytics, human and face sub-attribute recognition, all on the edge, while achieving near state-of-the-art accuracies. We divide up our analytics into four main components:

\subsection{Object Detection} 
Object Detection is a key component and starting point of our analytics pipeline. Tremendous progress is being achieved on this problem by the region-based family~\cite{girshick2015fast,ren2015faster} and the single-pass family i.e. YOLO~\cite{redmon2016you,yolov3} and SSD~\cite{SSD}. Even though the region based family (see Section \ref{subSec:advDLCV}) provides high detection accuracy, they prominently rely on 'Selective Search' for region proposals which hampers the detection speed. Even the fastest, highest accuracy region based detection algorithm, Faster R-CNN~\cite{ren2015faster} can achieve only 7 FPS, which is not a viable solution to problem scenarios that require real-time object detection. On the other hand, single-pass detectors like YOLO and SSD do not rely on bounding box proposals and still give significantly better results in terms of both speed and accuracy. Recently, ~\citeauthor{yolov3} released YOLOv3~\cite{yolov3}, which claims to be 3x faster than SSD with the same accuracy. Moreover, YOLO has a smaller derived network called Tiny YOLO which is capable of operating on a CPU. This paper builds upon the YOLO family of networks for performing object detection in real-time. Depending on the use case, we custom train the object detectors on the classes of objects relevant to the business needs of the deployment. For example, for monitoring safety in construction sites, we might care about objects such as humans, helmets, safety shoes etc. In these cases, we do not need to consider other objects such as cars, buses or bags etc. On the other hand, if it is a traffic scenario, the focus will be on vehicle classes, such as cars, trucks, buses, motorbikes etc.

\subsection{Face Detection and Recognition}
Another important part of our pipeline is face detection and recognition. For long, the detection framework laid out by~\citeauthor{ViolaJones}~\cite{ViolaJones} was the go-to for face detection. Though it was fast, it produced quite a lot of false positives. Another commonly used face detection algorithm was proposed by~\citeauthor{NPD}~\cite{NPD}. They extracted a feature from an image which was computed as the difference to sum ratio between two pixel values. They called it the 'Normalized Pixel Difference' (NPD)~\cite{NPD}. They used this feature along with a soft-cascade classifier to detect faces in the given image. The NPD face detector was fast and achieved state-of-the-art performance on srstandalrdts. However, we found that NPD was slow and required a lot of tuning to work in surveillance videos, due to inconsistent frame dimensions. As a result, we use a Single Shot Detector~\cite{SSD} model based on ResNet~\cite{chi2017end}. As illustrated in Section~\ref{facedetsubsec}, we see that the ResNet-SSD model outperforms NPD and Haar, both in terms of speed and accuracy. For face recognition, we use a ResNet~\cite{Resnet} based face embeddings trained by~\citeauthor{dlibFace}~\cite{dlibFace} on about three million images. In Section~\ref{table:faceRecognition}, we compare accuracy results of various deep and shallow face embeddings on a surveillance face recognition dataset.

\subsection{Sub-Attribute Recognition}
Based on the detected objects, we then perform sub-attribute recognition. The ability to recognize sub-attributes for a localized object from a larger image allows us to index an object in multiple ways. So, first we need to detect the position of the people by running the captured frames through an object detector, as discussed in Section~\ref{subSec:objDet}, after which we classify the object on the basis of its sub-attributes. 

On the localized people, we run human sub-attribute recognition, which consists of recognizing the age, gender, apparel type and color etc. Similarly, in the case of vehicles, we might be interested in the make and type of the vehicle. In the case of faces, we care about the recognition of face sub-attributes such as age, gender and emotion. In the case of other objects (e.g. bags, helmets etc.) we might be interested in properties like the color and size of the object. For tackling most of these sub-attribute recognition problems, we utilize two methodologies, viz. 'Transfer Learning' and 'Fine Tuning'.

\subsubsection{Transfer Learning}\label{sys:TransferLearning}
In this approach, we choose a pre-trained CNN model, ideally trained for a contextually similar problem. We then choose a layer of the model, which is used to extract the features of an image when it is forward passed through the network. This extraction of feature vectors is performed for all images in the training set. Upon the completion of the feature extraction, we train a multinomial logistic regression model. On the positive side, Transfer Learning allows us to quickly train models without a GPU. It generally achieves a high accuracy on a held-out test set, when trained on a small training set. However, Transfer Learning requires us to possess domain specific knowledge, and intricacies of the base CNN model to be used, in order to identify the feature extraction layer. Moreover, the identification of the layer involved quite some experimentation and heuristics.

\subsubsection{Fine Tuning}\label{sys:FineTuning}
While Fine Tuning a model, the original network architecture is modified to be compliant with our training set. The weights of the original network act as the base weights for the model, and we start its training as usual. This allows us to use the embeddings that the original network may have learned and build on top of them. Thus, the trained model is more robust and suited for our task. Unfortunately, the hyper parameters need to be tuned pertinently in order to obtain good results. Fine Tuning also requires a considerably large number of images than Transfer Learning. It also requires a GPU to train the network in a feasible amount of time.

\subsection{Tracking}
Finally, a very important piece of video analytics is tracking the detected objects and faces. Tracking is the process of locating the position of an entity across sequential frames in a video. In multi-object tracking, we are required to map the location of detected entities in a frame in the subsequent frames. Traditional position based algorithms fail when the detected entities are close to each other. We overcome this difficulty in our system by implementing the SORT algorithm~\cite{Bewley2016_sort}. The SORT algorithm is very fast and performs much better than position based tracking.

\section{Data Collection and Training Pipeline for Model Customization} \label{pipeline-overview}
Surveillance cameras have a fixed field of view and their orientations largely remain unchanged. Thus, in order to train highly accurate models, we obtain videos directly from the deployment locations. Videos, however, largely contain redundant data. Since each frame needs to be labeled by a human labeler, this will increase the cost of labeling. To tackle this issue, we collect a set of diverse frames by summarizing the video using submodular functions~\cite{NIPS2014_5415,wei2015submodularity,sahoo2017unified}. This drastically reduces the number of images that need to be annotated in order to train the model. Thus, decreasing the overall turnaround time without compromising on the accuracy of the model.
\subsection{Removing Redundant Frames via Diversity Models}
Given a set $V = \{1, 2, 3, \cdots, n\}$ of items which we also call the \emph{Ground Set}, define a utility function (set function) $f:2^V \rightarrow \mathbf{R}$, which measures how good a subset $X \subseteq V$ is. In our case, the ground set comprises of frames from the video sampled at a particular FPS (say 1 frame per second). A special class of set functions, called submodular functions, form very natural models for diversity. Submodular functions exhibit a property that intuitively formalizes the idea of ``diminishing returns''. That is, adding some instance $x$ to the set $A$ provides more gain in terms of the target function than adding $x$ to a larger set $A'$, where $A \subseteq A'$.  Informally, since $A'$ is a superset of $A$ and already contains more information, adding $x$ will not help as much. For more examples of submodular functions, see~\cite{NIPS2014_5415,wei2015submodularity,sahoo2017unified}. 

Given a budget (say of 500 frames from the video), we would like to choose the most diverse frames, based on a diversity model $f$, so as to ensure the best coverage of the entire video. Consider a simple greedy algorithm, which, starts with $X = \emptyset$ and iteratively adds an element $j \notin X$ which maximizes the gain $f(X \cup j) - f(X)$. We stop the greedy algorithm when the budget constraint is satisfied. One can show that this is a near optimal solution to the problem of maximizing the diversity model $f$ subject to a budget constraint. Given several videos from deployment locations, we summarize these videos to extract the most diverse frames, and then label the frames.

\subsection{Data Augmentation for Generalization}~\label{subSec:dataAugmentation}
Image classification tasks often have a very pertinent problem of a lack of sufficient labeled data, which is the most important commodity in solving a supervised learning problem. This hinders implementations of such models on a wide variety of real world problems. In addition to this, a model trained on a smaller training set is bound to over-fit the training set and will not generalize well.

As a solution to this, the idea of data augmentation is put forward, which in essence is to create more training samples from the information which is already present in the training set. Hence, by generating a larger training set, we counter the problem of over-fitting and also help the model to generalize better. Early implementations of successful data augmentation techniques can be seen on the MNIST dataset~\cite{lecun-mnisthandwrittendigit-2010}. 

Traditionally, the techniques which are applied as a part of data augmentation include rotation, flipping, shearing and changing the color of the image. These affine transformations follow the format of $y = Ax + B$, where $x$ is the representation of the original image, $A$ and $B$ are transformation parameters and $y$ is the representation of the transformed image.

In our approach, we implement an augmentation pipeline which performs these affine transformations on images by randomly choosing the transformation parameter values, hence creating a diverse set of new images. This pipeline generates a training set with an equal number of samples for each class, which is calculated as the average of the number of samples per class, pre-augmentation. To avoid adding extra noise to the dataset, transformation parameters are selected only belonging to a particular range, for example, the image rotation may not exceed more than 10 degrees.

\section{Video Analytics Results}
This section goes over the results for the different video analytics building blocks discussed above in various customer deployments. Due to shortage of space, we focus mainly on results for object detection, face recognition, human and face sub-attribute. The pattern of the results, however, hold for the other analytics as well not discussed in this paper. 

\subsection{Object Detection}\label{subSec:objDet}
In this paper, we use YOLO for illustrating the importance of custom training such networks, by providing a head-to-head comparison between off-the-shelf models trained on generic datasets and models trained on custom datasets. Also, we show that there is not a huge gap in performance between custom trained YOLO and Tiny YOLO models from a deployment perspective by providing a similar head-to-head comparison.

These state-of-the-art networks are extensively used to solve object detection problems in various scenarios like counting students in a classroom, detecting vehicles running on the highway, etc. These problems get more complicated when subcategories like boy/girl student for the first case and vehicle make/model/color for the second case are needed to be identified. Generally these models are trained on ImageNet~\cite{ILSVRC15}, PASCAL VOC~\cite{VOC}, Microsoft COCO~\cite{COCO} or any such standard dataset, these pre-trained models give good results when it comes to detecting objects in generic scenarios. ~\citeauthor{DBLP:journals/corr/HuangRSZKFFWSG016}~\cite{DBLP:journals/corr/HuangRSZKFFWSG016} provide a detailed comparison of these networks trained on the COCO dataset. However, they might not work well in all the real world scenarios due to the fact that the dataset used for training these models are very generic.

Below we describe four datasets that have been used throughout our experiments.
~\label{enum:dataset}
\begin{enumerate}
\item Classroom Dataset (736 images): 
This dataset consists of classroom images with varied seating arrangements, surroundings, class strength, etc. The main customer requirements for this deployment were getting accurate student counts, detecting whether class has started or not, and uniform compliance.
\item Community Center Dataset (5336 images): This dataset consists of indoor and outdoor images of a community place encompassing dense and sparse crowds of different age groups thereby providing an aggregated data for detecting the person class. The customer requirement was to get various statistics including counts, age/gender distribution, heat-map/flow-map etc.
\item Traffic Dataset (999 images): This dataset showcases running roads and highways consisting of various vehicles (car, bus, truck, bicycle, motorbike, three-wheelers), with different perspectives and densities.
\item Multi-Scenario Surveillance Dataset (8191 images): The Multi-Scenario Surveillance dataset is a blend of data from several customers, including the ones above. This dataset consists of the the most common classes seen in surveillance videos including persons, cars, buses, trucks, motorbikes etc. This dataset is similar to the PASCAL VOC dataset, except that it is focused on data from surveillance videos, instead of images downloaded from the Internet.
\end{enumerate}

Table \ref{table:YOLO mAP Comparisons} illustrates the performance of YOLO and Tiny YOLO models trained on the PASCAL VOC dataset, Multi-Scenario Surveillance dataset, and the above mentioned custom datasets in terms of their Mean Average Precision (mAP) at an Intersection over Union threshold (IoU) of 0.5, along with their inference times in milliseconds.

{\renewcommand{\arraystretch}{1.2}
\begin{table}
\caption{Comparison between models trained on PASCAL VOC, Multi-Scenario Surveillance (MSS) and Customized Datasets} % title of Table
\centering % used for centering table
\begin{tabular}{| p{2cm} | c | c | c | c | c | c | c |} % centered columns ( columns)
\hline
\multirow{4}{2cm}{\centering Target Dataset} & \multirow{4}{*}{Train} & \multicolumn{3}{c |}{YOLO} & \multicolumn{3}{c |}{Tiny YOLO} \\
\cline{3-8}
 &  & \multirow{3}{*}{mAP} & \multicolumn{2}{c |}{Inference Time} & \multirow{3}{*}{mAP} & \multicolumn{2}{c |}{Inference Time} \\
 & & & \multicolumn{2}{c |}{(in ms)} &  & \multicolumn{2}{c |}{(in ms)} \\
\cline{4-5}
\cline{7-8}
 &  &  & GPU & CPU &  & GPU & CPU \\
 \hline
\multirow{3}{2cm}{\centering Classroom} & VOC & 9.04 & 53 & 371 & 4.6 & 11 & 60 \\
\cline{2-8}
 & MSS & 58.79 & 43 & 310 & 36.7 & 14 & 98  \\ 
 \cline{2-8}
 & Custom & \textbf{66.16} & 43 & 314 & \textbf{54.3} & 14 & 96  \\ 
\hline
\multirow{3}{2cm}{\centering Community Center} & VOC & 40.9 & 34 & 276 & 26.8 & 10 & 80  \\
\cline{2-8}
 & MSS & 75.2 & 26 & 343 & 59.8 & 11 & 60  \\
 \cline{2-8}
 & Custom & \textbf{80.72} & 25 & 343 & \textbf{70.36} & 12 & 61  \\
 \hline
\multirow{3}{2cm}{\centering Traffic} & VOC & 18.29 & 14 & 271 & 16.61 & 5 & 45  \\
\cline{2-8}
 & MSS & 59.1 & 12 & 110 & 40.10 & 5 & 30  \\
 \cline{2-8}
 & Custom & \textbf{71.45} & 10 & 106 & \textbf{72.54} & 5 & 31  \\
 \hline
\multirow{2}{2.2cm}{\centering Multi-Scenario Surveillance} & VOC & 9.10 & 55 & 410 & 4.28 & 16 & 84  \\
\cline{2-8}
 & Custom & \textbf{47.22} & 51 & 334 & \textbf{32.47} & 6 & 27  \\
\hline %inserts single line
\end{tabular}
\label{table:YOLO mAP Comparisons} % is used to refer this table in the text
\end{table}
}
The following are the main takeaways from the results: 
\begin{enumerate}
\item In all cases, we see that the customized models perform better on held-out test sets compared to the off-the-shelf PASCAL VOC and Multi-Scenario Surveillance models, even though they are trained only with a fraction of the data.
\item As expected, the Multi-Scenario Surveillance dataset performs much better compared to the PASCAL VOC dataset. This is expected, since the Multi-Scenario Surveillance dataset consists of surveillance images, rather than images downloaded from the Internet.
\item Even though a custom trained YOLO model performs the best, the CPU latency is too high to run it in real-time. On the other hand, a custom trained Tiny YOLO model performs well from an accuracy perspective and yet takes about one fifth the time for inference compared to YOLO (on CPU).
\end{enumerate}

\subsection{Face Detection and Recognition}
% Pratik to write
Face detection and recognition in surveillance videos has long been of utmost importance. However, in todays world we are required to identify a lot more fine-grained attributes, such as age and gender, from a person's face. To be able to recognize these sub-attributes, it is necessary for us to obtain certain discerning features from a face in order to classify it. For this purpose, we use a CNN for extracting facial features by passing the image of a detected face through the model.

\subsubsection{Face Detection}\label{facedetsubsec}
Table~\ref{table:faceDetection} compares the face detection accuracy for Haar cascades, NPD and ResNet detector. The ResNet-SSD model wins from both, the speed and accuracy, perspectives on the FERET dataset. The timings for NPD and Haar cascade are obtained without tuning any hyper parameters such as the minimum and maximum face size. Though tuning these parameters may decrease the inference time (and make them comparable to SSD), they are very difficult to calibrate as they are circumstantially variant.
{\renewcommand{\arraystretch}{1.2}
\begin{table}
\caption{Detection Accuracy and Average Inference Times on the FERET Dataset}
\centering
\begin{tabular}{ | c | c | c | }
\hline
\multirow{2}{3cm}{\centering Detection Algorithm} & \multirow{2}{3cm}{\centering Precision (in \%)} & \multirow{2}{3.5cm}{\centering Average Inference Time (in ms)} \\
 &  &  \\
\hline
\centering Viola-Jones Haar Cascade & 53.39 & 114.94 \\
\hline
\centering NPD Detector & 73.03 & 148.61 \\
\hline
\centering ResNet-SSD &  \textbf{97.81} &  \textbf{60.29} \\
\hline
\end{tabular}
\label{table:faceDetection}
\end{table}
}
\subsubsection{Face Recognition}\label{subSec:faceRec}
In order to perform face recognition on a detected face, we first need to train a model on the set of faces which may be detected. As we would only have a few images of every distinct person, we chose the Transfer Learning approach, elaborated in Section~\ref{sys:TransferLearning}, to train our face recognition models. We extracted features from multiple pre-trained CNNs, namely Deep Face~\cite{Parkhi15}, DLib-ResNet~\cite{dlibFace} and OpenFace~\cite{Baltrusaitis2016}, and compared the accuracy of the model on the FERET~\cite{FERET} and Community Center datasets, which are illustrated in Table~\ref{table:faceRecognition}.

{\renewcommand{\arraystretch}{1.2}
\begin{table}
\caption{Face Recognition Accuracies on the FERET~\cite{FERET} and Community Center Datasets}
\centering
\begin{tabular}{ | c | c | c | }
\hline
\multirow{2}{*}{\centering Dataset} & \multirow{2}{*}{\centering CNN Model Used} & \multirow{2}{*}{\centering Recognition Accuracy (in \%)} \\
 &  &  \\
\hline
\multirow{3}{*}{\centering FERET} & \centering Deep Face Recognition & 95 \\
\cline{2-3}
 & \centering DLib-ResNet &  \textbf{99.76} \\
 \cline{2-3}
 & \centering OpenFace & 77.52 \\
\hline
\multirow{3}{*}{\centering Community Center} & \centering Deep Face Recognition & 92.55 \\
\cline{2-3}
 & \centering DLib-ResNet &  \textbf{92.62} \\
 \cline{2-3}
 & \centering OpenFace & 68.70 \\
\hline
\end{tabular}
\label{table:faceRecognition}
\end{table}
}

\subsection{Human Sub-Attribute}
For our experimentation, we classify the detected people on the basis of four sub-attributes: Full Body Age, Full Body Gender, Upper Body Apparel Type, and Upper Body Apparel Color.

We use the aforementioned techniques of transfer learning and fine-tuning to establish a baseline using out-of-the-box models trained on the PETA dataset. We progressively demonstrate the use and results of techniques such as data augmentation, transfer learning and fine-tuning on customized models in the following sections.

\subsubsection{Off-the-shelf model}
For our baseline, we use transfer learning and fine-tuning on AlexNet and GoogLeNet, trained on the PETA dataset~\cite{PETA}. The PETA dataset contains a wide variety of generic surveillance cases. We modify the dataset to only use the classes which were the most common. The model accuracy is evaluated over a held-out test set. The results obtained are given in Table~\ref{table:baselineTL}. 
{\renewcommand{\arraystretch}{1.2}
\begin{table}
\caption{Accuracy Results for Human Sub-Attribute Recognition (in \%)}
\centering
\begin{tabular}{ | c | c | c | c | c | c |}
\hline
\multirow{2}{2cm}{\centering Training Approach} & \multirow{2}{*}{\centering CNN Model} & \multirow{2}{*}{\centering Age} & \multirow{2}{*}{\centering Gender} & \multirow{2}{1.5cm}{\centering Apparel Type} & \multirow{2}{1.5cm}{\centering Apparel Color} \\
 &  &  &  &  & \\
\hline
\multirow{2}{2cm}{\centering Transfer Learning} & AlexNet & 67.2 & 75.1 & 59.8 & 53 \\
\cline{2-6}
& GoogLeNet & 65.8 & 76.7 &  \textbf{64.1} & 52.5 \\
\hline
Fine Tuning & AlexNet &  \textbf{77.4} &  \textbf{83.8} & 63 &  \textbf{65.24} \\
\hline
\end{tabular}
\label{table:baselineTL}
\end{table}
}
For simplicity, we only look at AlexNet and GoogleNet models. From Table~\ref{table:baselineTL}, we observe that the fine-tuned AlexNet seems to perform the best on nearly all four cases. Since the generic dataset is quite big (consisting of 10-20k images per class), it is expected that the fine-tuned model works better than the transfer learning, as transfer learning is essentially just learning the last layer, whereas fine-tuning is fitting the entire CNN to the data.

\begin{wrapfigure}{R}{0.5\textwidth}
\centering
\includegraphics[width=0.5\textwidth]{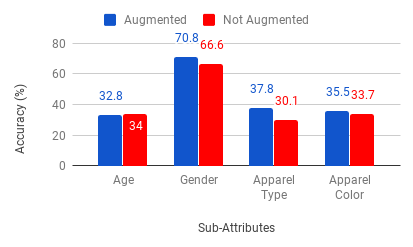}
\caption{Test Accuracies (in \%) on the Scraped Image Set.}
\label{generic_augmentation}
\end{wrapfigure}

\subsubsection{Data Augmentation}

The first enhancement we can make to this off-the-shelf model is to augment the training set in each case. The motivation behind augmentation is that we want the representation of each class in the dataset for any given task to be equal. So, using the data augmentation techniques mentioned in Section~\ref{subSec:dataAugmentation}, we create a new training set from the training set used of the PETA Dataset. Since, the original test set was created as a 20\% fraction of the dataset, there is a bias towards certain classes in the test set. But, the augmented training set has the same sample size for each class. Thus, in order to avoid distribution bias in our evaluation, we create a separate test set, called the 'Scraped Image Set' which contains images scraped from the web and labeled manually. From Figure~\ref{generic_augmentation}, we can see that a model with an augmented training set is more suited to generalization as opposed to one without it. The lack of difference in the age classification case can be explained by the fact that the original dataset has a nearly equal distribution amongst its classes and augmentation did not make a significant difference to the training set.

\subsubsection{Model Customization}
In this section, we compare a custom trained model against the off-the-shelf models generated in the above section. Towards this end, we perform transfer learning and fine-tuning on the dataset consisting of images from the deployment scenario. This mitigates the problem of non-inclusion of attributes for diverse cases. For example, generic datasets such as the PETA dataset contains classes for clothing such as Coat, Suit, Shirt, etc. which would be a problem if deployed in a place with a different clothing norm such as Asia.

\begin{figure}
  \centering
  \begin{minipage}[b]{0.4\textwidth}
    \includegraphics[width=\textwidth]{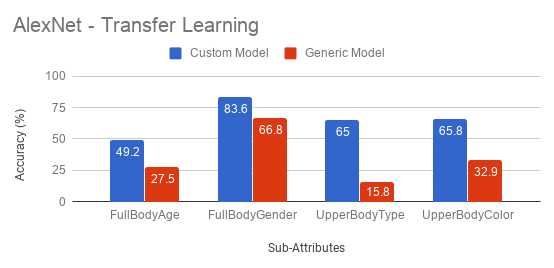}
  \end{minipage}
  \hfill
  \begin{minipage}[b]{0.4\textwidth}
    \includegraphics[width=\textwidth]{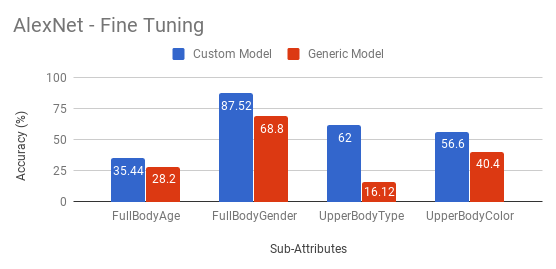}
  \end{minipage}
  \caption{Accuracies of models trained on the Community Center dataset}
  \label{customization}
\end{figure}

For our experiments, we used the Community Center dataset described in Section~\ref{enum:dataset}. We compare the accuracies of the off-the-shelf model against customized models on the test set of the Community Center dataset. For ease of experimentation, we only perform the transfer learning and fine-tuning on AlexNet. The results can be seen in Figure~\ref{customization}. We see that both in the case of the Transfer learned and fine-tuned models, the customized models perform better than the generic one.

The following are the insights from the results:
\begin{enumerate}
\item Generally, fine-tuning a model works better than transfer learning. However, if the training dataset is small, fine-tuned models tend to over-fit and need more careful hyper-parameter tuning. This is costly both in terms of resources used and time spent. As it can be seen from Figure~\ref{customization}, the performance of transfer learning is just as good if not better than fine-tuning. It is much faster to perform transfer learning and works better on fewer data. Hence, in a custom deployment scenario, where procuring a large custom dataset might be difficult, the use of transfer learning on custom datasets is recommended.
\item In cases where multiple sub-attributes need to be recognized, loading fine-tuned models is resource expensive, since we load multiple models in memory. In such scenarios, transfer learning is beneficial as a single base CNN can be used as a feature extractor for multiple sub-attribute models.
\item Data augmentation can be used, not only to increase the size of the training set, but also to increase the generalizability of the models. Due to space constraints, the comparison between augmented and non-augmented custom trained models could not be shown, but we observe similar patterns.
\item Even with a relatively small amount of data, performing transfer learning on a custom model is better than using a generic model. So, a customized model should be used over a generic model whenever possible.
\end{enumerate}

\subsection{Face Sub-Attribute}\label{subSec:faceSubAttr}
Recognizing fine-grained attributes such as a person's age or gender, from their face, has become an important task in recent times. To tackle this problem, we use our Transfer Learning approach to train a multinomial logistic regression model on the localized faces of people. Similar to the human sub-attribute recognition, we compare customized transfer learned models against off-the-shelf models for age and gender. As off-the-shelf models, we directly use the Deep Expectation (DEX) models~\cite{Rothe-IJCV-2016} and, the AgeNet and GenderNet model by~\citeauthor{LH:CVPRw15:age}~\cite{LH:CVPRw15:age}. Both these models have been trained to recognize the age and gender of a detected face. For customization, we use transfer learning on the Deep Face Recognition model~\cite{Parkhi15}, along with the AgeNet, GenderNet, DEX-Age and DEX-Gender. All the results are obtained on the Community Center dataset.

In Table~\ref{table:baselineFaceSub}, we observe that the off-the-shelf models generalize poorly on the Community Center dataset. Since, these off-the-shelf models are not customized for faces detected from surveillance cameras. Using the same Transfer Learning approach mentioned in the human sub-attribute case, we train multinomial logistic regression models on the different base CNNs discussed above.
{\renewcommand{\arraystretch}{1.2}
\begin{table}
\caption{Off-the-shelf and Customized Age and Gender Classification Results on the Community Center Dataset}
\centering
\begin{tabular}{ | c | c | c | }
\hline
\multirow{2}{*}{\centering Category} & \multirow{2}{*}{\centering CNN Model} & \multirow{2}{*}{\centering Classification Accuracy (in \%)} \\
 &  &  \\
\hline
\multirow{2}{*}{\centering Age Generic} & \centering DEX-Age & 28.03 \\
\cline{2-3}
 & \centering AgeNet &  26.43 \\
\hline
\multirow{5}{*}{\centering Age Customized} & \centering DEX-Age & 73.57 \\
\cline{2-3}
 & \centering DEX-Gender &  68.11 \\
 \cline{2-3}
 & \centering AgeNet &  63.86 \\
 \cline{2-3}
 & \centering GenderNet &  62.09 \\
 \cline{2-3}
 & \centering Deep Face Recognition &  \textbf{91.00} \\
\hline
\multirow{2}{*}{\centering Gender Generic} & \centering DEX-Gender & 72.08 \\
\cline{2-3}
 & \centering GenderNet &  75.00 \\
\hline
\multirow{5}{*}{\centering Gender Customized} & \centering DEX-Age & 91.24 \\
\cline{2-3}
 & \centering DEX-Gender &  92.52 \\
 \cline{2-3}
 & \centering AgeNet &  72.08 \\
  \cline{2-3}
 & \centering GenderNet &  83.21 \\
 \cline{2-3}
 & \centering Deep Face Recognition &  \textbf{97.26} \\
\hline
\end{tabular}
\label{table:baselineFaceSub}
\end{table}
}

Firstly, we observe that the transfer learned models have considerably superior accuracies for both age and gender recognition problems. The Deep Face Recognition models perform the best on both tasks. It is worth noting that the Deep Face Recognition model learns embeddings to distinguish between people. Thus, it possesses a better embeddings representation than the other models for the age and gender recognition tasks. Moreover, we should also note that the age models seem to perform better on age recognition (compared to the gender model) and their gender counterparts perform better on gender recognition (compared to the age model). This also matches our intuitions.
 
\section{Conclusions and Lessons learned}
This paper provides an overview of what it takes to build an end-to-end video analytics system which is capable of performing deep learning in real-time on a CPU. We go over the data collection tricks and training pipelines for fast experimental turn around. We also share the significant amount of practical experience we have gained by deploying models at customer locations, including tricks like where to use data augmentation, the effectiveness of transfer learning vs fine-tuning, and the amount of data required for custom training.
\\
The major takeaways from this paper are as follows:
\begin{itemize}
\item Deep Learning does not necessarily require GPU cloud servers. It is possible to get high accuracy using on-premise CPU deployments.
\item Customization will always give better results than off-the-shelf models.
\item Customization provides superior accuracies compared to off-the-shelf models even with fraction of the training dataset.
\item Tricks such as data summarization and data augmentation in our customized training pipeline ensures we obtain high accuracy with a smaller training set.
\end{itemize}

\bibliographystyle{abbrvnat}
\bibliography{ecml}
\end{document}